# Classification of social media Toxic comments using Machine learning models

K.Poojitha , A.Sai Charish , M.Arun Kumar Reddy , S Ayyasamy

**ABSTRACT**

Toxic comments are disrespectful, toxic, abusive and unreasonable that makes the users leaves the discussion. In present generation, social media has become major part of everyone's life. People are getting bullied for numerous reasons. Not all people on internet are interested in participating nicely, some will vent their anger, insecurities and prejudices. This anti-social behaviour often occurs during the debates in comment section, discussion and fights often takes place in the online platform where it involves rude and disrespectful comments which are known are toxic comments. Comments containing explicit language can be classified into myriad categories such as Toxic, Severe Toxic, Obscene, Threat, Insult, and Identity Hate. The threat of abuse and harassment means that many people stop expressing themselves and give up on seeking different opinions. To protect users from being exposed to offensive language on online forums or social media sites, companies have started flagging comments and blocking users who are found guilty of using unpleasant language. Several Machine Learning models have been developed and deployed to filter out the unruly language and protect internet users from becoming victims of online harassment and cyberbullying We will aim to create a classifier which classifies the comments between the toxic and non-toxic comments which helps the organizations to get the better picture of the comment section to examine the toxicity with high accuracy using Lstm-cnn model

## I. INTRODUCTION

Social media is a place where a lot of discussions happen, being anonymous while doing so has given the freedom to many people to express their opinions freely. But people who disagree with a point of view extremely can misuse this freedom sometimes. Sharing things that you care about will become a difficult task with this constant threat of harassment or toxic comments online. This will eventually lead to people not sharing their ideas online and stop asking for other people's opinion on them. Unfortunately, the social media platforms face these issues all the time and find it difficult to identify and stop these toxic remarks before it leads to the abrupt end of conversations.

In this we will be using Natural Language Processing with Deep neural networks to solve this problem of identifying the toxicity of online comments. Word embeddings will be used in conjunction with recurrent neural networks with Long Short Term Memory (LSTM), Convolutional Neural Networks (CNN), and separately and see which model fits and works best.

Text classification has become one of the most useful applications of Deep Learning, this process includes techniques like Tokenizing, Stemming, and Embedding. This paper uses these techniques along with few algorithms, that are used to classify online comments based on their level of toxicity.

We proposed a neural network model to classify the comments and compared the model's accuracy with some other models like Long Short Term Memory (LSTM) and Convolutional Neural Network .The comments are first passed to a tokenizer or vectorizer to create a dictionary of words, then an embedding matrix is created after which it is passed to a model to classify.

## II. LITERATURE REVIEW

Toxic comments on social media platforms have been a source of a great stir between individuals and groups. A toxic comment is not only verbal violence but includes the comment that is rude, disrespectful, negative online behaviour, or other similar attitudes that make someone leavea discussion. Therefore, the toxic comments identification on social platforms is an important task that can help to maintain its interruption and hatred-free operations. Consequently, a large variety of toxic comment approaches have been proposed. Three characteristics concerning toxic classification are evaluated: classification, feature dimension reduction, and feature importance.

The author of "Impact of SMOTE on Imbalanced Text Features for Toxic Comments Classification using RVVC Model" discusses about the ensemble approach called regression vector voting classifier(RVVC), to identify the toxic comments over the social media platforms. The dataset for this is taken from Kaggle which is a multi-label dataset which contains labels as toxic, severe_toxic, threat, insult, and identity_hate. The values of the dataset are given as binary values which contains 158,640 comments in total with toxic comments. Owing to the study of higher accuracy ensemble model their experiment indicated the good performance and they combined the LR and SVC model to get the higher accuracy which is known as RVVC. They conducted the experiments using originally imbalanced dataset with TF-IDF and BoW separately. The results proved that RVVC outperforms all other individual models when TF-IDF features are used with SMOTE balanced dataset and achieves an accuracy of 0.97. Despite the better performance of proposed model, its computational complexity is higher than individual models. The authors for "Toxic Comment Classification Implementing CNN combining Word Embedding Technique" says that Despite this model, it only classifies into six labels which further improves to another model where it first classifies whether the comment is positive or negative and then classifies into six labels. The dataset used is Wikipedia's talk page edits, collected from Kaggle. They proposed the ensembled model called Convolution Neural Network(CNN) which is structured as data cleaning, adopting the NLP techniques, stemming, and converted word into vector by word embedding techniques. The accuracy for this model is calculated based on ROC – AUC. The receiver operating characteristic curve (ROC) score is 98.46% and the area under score (AUC) score for this model is 98.05% which is much accurate than the previous model and the existing works. There's another technique where it uses the deep learning model recurrent neural networks(RNN) to perform text classification on a multi-label text dataset to identify different forms of internet toxicity. The paper "Application of Recurrent Neural Networks in Toxic Comment Classification" discusses about this aspect. The dataset used for this methodology is the public dataset which is provided by Conversation AI team. Even Though the previous models has given the accurate toxicity scores, the models still miss classify some texts that share similar patterns as toxic comments which can be reduced using the RNN method. They used the Word2Vec embedding model to train the model to remove the noise in the data and the methodologies used are the recurrent neural network and done the comparison analysis with GRU. They have successfully employed the word2vec embedding and recurrent neural network in building a toxic comment classification model and achieved high accuracy with low cost. Even with the existing models have achieved high accuracy, building the model takes more time and complexity of the model will higher. This leads to Automatic detection. The paper "Automatic toxic Comment Detection Using DNN" discusses about the automatic tools which where build using the LSTM and RNN to improve the accuracy and provide the results much faster than the traditional methods. This paper compared three state-of-art unsupervised word embedding models which were Mikolov's word embedding, fastText subword embedding, BERT Wordpiece model. The dataset used is Wikipedia Detox Corpus which talks about English Wikipedia talk pages. All the models were compared against the BERT fine-tuning. And on experiments and results, it showed that BERT fine-tuning is the most efficient model at this automatic toxic classification task. This can further developed into the speech toxic recognition using more advanced models like XLNet model. There's another model where the hybrid models are produced to get the accurate score much more than CNN model. The paper "Toxic Comment Classification Using Hybrid Deep Learning Model" discusses about the hybrid models used which are Bidirectional gated recurrent network, convolution neural network and they achieved the accuracy of 98.39% and the f1 score of this model was 79.91%. As much as toxic comment classification is important, toxic span prediction also play the similar role which helps to build more automated moderation systems. The paper "Multi-task learning for toxic comment classification and rationale extraction" discusses about the multi-task learning model using the Toxic XLMR for bidirectional contextual embeddings of input text for toxic comment classification and a Bi-LSTM CRF layer for toxic span and rationale identification. The dataset used was curated from Jigsaw and Toxic span prediction dataset. The model has outperformed the single task models on the curated and toxic span prediction models by 4% and 2% improvement for classification. The future improvements can be added to take more delicate context and handle the subtle differences in usage of keywords. The paper "Vulgarity Classification in comments using SVM and LSTM" discusses about the hybrid model of SVM and RNN-LSTM. The data is vectorized using TF-IDF and bag of words. This paper also discusses the nature of the dataset. The results found to give a promising assurance in finding a solution. The paper "Modern Approaches to Detecting and classifying Toxic Comments using Neural Networks" discusses about the algorithms constructed using deep learning technologies and neural networks that solve the problem of detecting and classifying toxic comments. The algorithms are tested and trained on a large training set and tagged by Google and Jigsaw which was taken from Kaggle.

### III. MATERIALS AND METHODS

This study uses different techniques, methods, and tools for the classification of toxic and non-toxic comments. Also, various preprocessing steps, data re-sampling methods, features extraction techniques, and supervised machine learning models are adopted for the said task.

## A. DATA DESCRIPTION

This study aims at the automatic classification of toxic and non-toxic comments from social media platforms. Various machine learning models are utilized for this purpose to evaluate their strength for the said task. For evaluation, the selected models are trained and tested with binary class datasets. Traditionally, toxic comments are grouped under several classes such as hate, toxic, threat, severe toxic, obscene, insult and non-toxic, etc. We follow a different approach by grouping the comments under two classes, toxic and non-toxic. The original dataset which is taken from Kaggle [30], is a multi-label dataset and contains labels such as toxic, severe_toxic, obscene, threat, insult, and identity_hate. The non-toxic comments belong to one class, while from the other comments only those comments are selected that have toxic labels. It means that the comments that label severe_toxic, obscene, threat, insult, and identity_hate are not selected. For example, Table 1 shows that 'comment2' is only toxic and 'comment 3' is non-toxic. For our experiment, both 'comment 2' and 'comment 1' are selected under toxic and no-toxic classes, but 'comment 1' and 'comment 4' are not select.

TABLE 1. Example of various classes in the original dataset.

| Comment_text | Comment 1 | Comment 2 | Comment 3 | Comment 4 |
|---|---|---|---|---|
| identity_hate | 0.0 | 0.0 | 0.0 | 0.0 |
| insult | 1.0 | 0.0 | 0.0 | 0.0 |
| obscene | 1.0 | 0.0 | 0.0 | 0.0 |
| severe_toxic | 0.0 | 0.0 | 0.0 | 0.0 |
| threat | 0.0 | 0.0 | 0.0 | 0.0 |
| toxic | 0.0 | 1.0 | 0.0 | 0.0 |
| toxicity | 0.0 | 0.0 | 0.0 | 1.0 |

We extract only toxic and non-toxic comments from the dataset and Table 2 shows the ratio of toxic and non-toxic comments in the dataset used for experiments. The ratio of toxic and non-toxic comments in the dataset is not equal which shows the imbalanced data problem. The performance of the classifiers could be affected due to an imbalanced dataset.

TABLE 2. Number of records for toxic and non-toxic comments.

| Category | No. of comments | Experimental Data |
|---|---|---|
| Non-Toxic | 143346 | 70000 |
| Toxic | 15294 | 15294 |
| Total | 158640 | 85294 |

The dataset contains 158,640 comments in total with toxic comments having the lowest ratio in the dataset, i.e., 15,294 while non-toxic comments are 143,346. It makes a huge difference and makes the dataset highly imbalanced. Due to the large size of the dataset, only 70,000 non-toxic comments are randomly selected for the experiments.

## B. PREPROCESSING STEPS

Pre-processing techniques are applied to clean the data which helps to improve the learning efficiency of machine learning models [31]. For this purpose, the following steps are executed in the given sequence.

**Tokenization:** is a process of dividing a text into smaller units called 'tokens'. A token can be a number, word, or any type of symbol that contains all the important information about the data without conceding its security.

**Punctuation removal:** involves removing the punctuation from comments using natural language processing techniques. Punctuations are the symbols that are utilized in sentences/comments to make the sentence clear and readable for humans. However, it creates problems in the learning process of machine learning algorithms and needs to be removed to improve their learning process. Some common punctuation marks are mostly used such that colon, question marks, comma, semicolon, full-stop/period, etc. ?:,;.[]() [32].

**Number removal:** is also a part of preprocessing which helps to improve the performance of the machine learning algorithms. Numbers are unnecessary and do not contribute to the learning of text analysis approaches. Removing the numbers increases the efficiency of models and decreases the complexity of the data.

**Stemming:** is an important part of preprocessing because it increases the performance by clarifying affixes from sentences/comments and converting the comments into the original form. Stemming is the process of transforming a word into its root form. For example, different words have the same meaning such as: 'plays', 'playing', 'played' are modified forms of 'play'. Stemming is implemented using the Porter stemmer algorithms [33].

**Spelling correction:** is the process of correcting the misspelled words. In this phase, the spelling checker is used to check the misspelled words and replace them with the correct word. Python library 'pyspellchecker' provides the necessary features to check the misspelled words and is used for the experiments [34].

**Stopwords removal:** Stopwords are those English words that do not add any meaning to a sentence. So these can be removed by stopwords removal without affecting the meaning of a sentence. The removal of stop-words increases the model's performances and decreases the complexity of input features [35].

## C. FEATURE ENGINEERING

Feature engineering aims at discovering useful data features or constructing features from original features to train machine learning algorithms effectively [36]. The study [37] concludes that feature engineering can improve the efficiency of machine learning algorithms. 'Garbage out' is a corporate proverb used in machine learning which implies that senseless data used as the input, yields meaningless output. In contrast, more information-driven data will yield favorable results. Hence, feature engineering can derive useful features from raw data which helps to improve the reliability and accurateness of learning algorithms. In the proposed methodology, two feature engineering methods are used including the bag of words and term frequency-inverse document frequency.

### D. BAG-OF-WORDS

The bag of words (BoW) technique is used to extract features from the text data. The boW is easy to implement and understand besides being the simplest method to extract features from the text data. The boW is very suitable and useful for language modeling and text classification. The 'CountVectorizer' library is used to implement BoW. CountVectorizer calculates the occurrence of words and constructs a spare database matrix of words [38]. The boW is a pool of words or features, where every feature is categorized as a label that signifies the occurrences of the categorized feature.

### E. TERM FREQUENCY-INVERSE DOCUMENT FREQUENCY

TF stands for term frequency and IDF stands for inverse document frequency of the word. The TF-IDF is a statistical analysis that is used to determine how many relevant words are in a list or corpus. The value increases with the number of times a word is shown in the text but is normalized by the word occurrence in the document [39].

- Term Frequency (TF): is the frequency of a term given in the text of a document. Because each document is dissimilar in size, it is likely that in long documents a word will occur more often than the shorter ones. To normalize, the term frequency is also divided by the length of the text.

$$F(t) = \frac{\text{No. of times } t \text{ appears in a document}}{\text{Total no. of terms in the document}} \quad (1)$$

- Inverse Document Frequency (IDF): is a rating of how infrequent the term is in a given document. IDF indicates the importance of a word on account of its rareness. The rare words have a higher IDF score.

$$IDF(t) = \log_e \frac{\text{(Total no. of documents)}}{\text{(No. of documents with term } t \text{ in it)}} \quad (2)$$

TF-IDF is then calculated using both TF and IDF using

$$TF - IDF = TF_{t,d} * \log N, \quad (3) \, D_f$$

where the $TF_{t,d}$ is frequency of term $t$ in document $d$.

### F. DATA RE-SAMPLING TECHNIQUES

Data re-sampling techniques are used to solve imbalanced dataset problems. The imbalanced dataset contains an unequal ratio of the target classes and can cause problems in classification tasks because models can over-fit on the majority class [40]. To solve this problem different data re-sampling techniques have been presented. In this study, two types of re-sampling techniques are used including under-sampling and over-sampling.

#### 1) RANDOM UNDER-SAMPLING

Under-sampling reduce the size of the dataset by deleting example of the majority class. For the under-sampling, a random under-sampling approach is used in the current study. In the random under-sampling, the major class examples are rejected at random and deleted to balance the distribution of the target classes. Simply we can say that under-sampling aims to balance class distribution by randomly deleting majority class examples. The random under-sampling technique is one of the widely used re-sampling approaches and selected due to its reported performance [41]–[44].

#### 2) SYNTHETIC MINORITY OVER-SAMPLING TECHNIQUE

Over-sampling is a technique in which the number of samples of the minority class is increased in the ratio of the majority class. Over-sampling increases the size of data which generates more features for model training and could be helpful to increase the accuracy of the model. In this study synthetic minority over-sampling technique (SMOTE) is used for over-sampling. SMOTE is a state-of-art technique that was proposed in [45] to solve the overfitting problem for imbalanced datasets. SMOTE randomly picks up the smaller class and finds the K-nearest neighbors of each smaller class. The picked samples are evaluated using the K-nearest neighbor for that particular point to construct a new minority class. SMOTE is adopted on account of the results reported in [46]–[49].

### G. PROPOSED METHODOLOGY

Ensemble learning is widely used to attain high accuracy for classification tasks. The combination of various models can perform well as compared to individual models. Owing to the high accuracy of ensemble models, this study leverage an ensemble model to perform toxic comments classification. Our experiments indicate the good performance from Lstm and CNN, so to further improve the performance, this study combines these models. The proposed approach is called (Lstm-Cnn). it ensure that the class with a high predicted probability by two classifiers will be considered as the final prediction.

Algorithm 1 shows the working of the proposed Lstm-cnn models and explains how it combines the Lstm and CNN for toxic comment classification. Let Lstm and CNN be the two models and 'toxic' and 'non-toxic' be the two classes, then the prediction can be made using the following equation

$$\text{Lstm-Cnn} = \text{argmax}\{Toxic_{prob}, NonToxic_{prob}\} \quad (4)$$

**Algorithm 1** Algorithm for Toxic Comments Classification

**Input:** Corpus-text comments
**Output:** Class-Toxic and what type of toxic or Non-Toxic

1: TLstm → Trained Lstm
2: TCnn → Trained cnn
3: **for** *i* in Corpus **do**
4: $\quad ToxicPob_{LR} \rightarrow Tlstm(i)$
5: $\quad NonToxicPob_{LR} \rightarrow Tlstm(i)$
6: $\quad ToxicPob_{SVC} \rightarrow Tcnn(i)$
7: $\quad NonToxicPob_{SVC} \rightarrow Tcnn(i)$
8: $\quad \text{Lstm-Cnn}_{Pred} \rightarrow \text{argmax}((ToxicPob_{LR} + ToxicPob_{SVC})/2, (NonToxicPob_{LR} + NonToxicPob_{SVC})/2)$
9: **end for**
10: $Toxic|Non-Toxic \rightarrow$ Lstm-Cnn prediction

where argmax is used in machine learning for finding the class with the largest predicted probability. The $Toxic_{prob}$ and $NonToxic_{prob}$ indicate the joint probability of toxic and non-toxic classes by the Lstm and cnn models and are calculated as follows

$$Toxic_{prob} = \frac{ToxicProb_{Lstm} + ToxicProb_{cnn}}{2} \quad (5)$$

$$NonToxic_{prob} = \frac{NonToxicProb_{Lstm} + NonToxicProb_{cnn}}{2} \quad (6)$$

where $ToxicProb_{lstm}$, and $ToxicProb_{cnn}$ are the probability for toxic class by Lstm and Cnn, respectively while $NonToxicProb_{Lstm}$, and $NonToxicProb_{SVC}$ are the probability scores for the non-toxic class by Lstm and cnn, respectively.

To illustrate the working of the proposed Lstm-Cnn model, the values for one sample are taken from the dataset used for the experiments. Lstm and cnn given probabilities for the sample data are

- $ToxicProb_{Lstm}$ = 0.6
- $NonToxicProb_{cnn}$ = 0.4
- $ToxicrPob_{lstm}$ = 0.5
- $NonToxicProb_{cnn}$ = 0.5

The combined $Toxic_{prob}$ and $NonToxic_{prob}$ are calculated as follows

$$Toxic_{prob} = \frac{0.6 + 0.5}{2} \quad (7)$$

$$NonToxic_{prob} = \frac{0.4 + 0.5}{2} \quad (8)$$

Then argmax function is applied to select the class with the higher probability. Here the largest prediction probability is for the Toxic class so the final prediction by Lstm-Cnn will be Toxic class.

$$RVVC = \text{argmax}\{0.55, 0.45\} \quad (9)$$

The flow of the proposed methodology is shown in Figure 1. In the proposed methodology, the toxic comment classification problem is solved using Lstm and cnn. For classification, the dataset is obtained from youtube [30] which contains toxic comments. Several preprocessing steps are carried out on the dataset to clean the data. After data cleaning, two feature extraction approaches including TF-IDF and are applied.

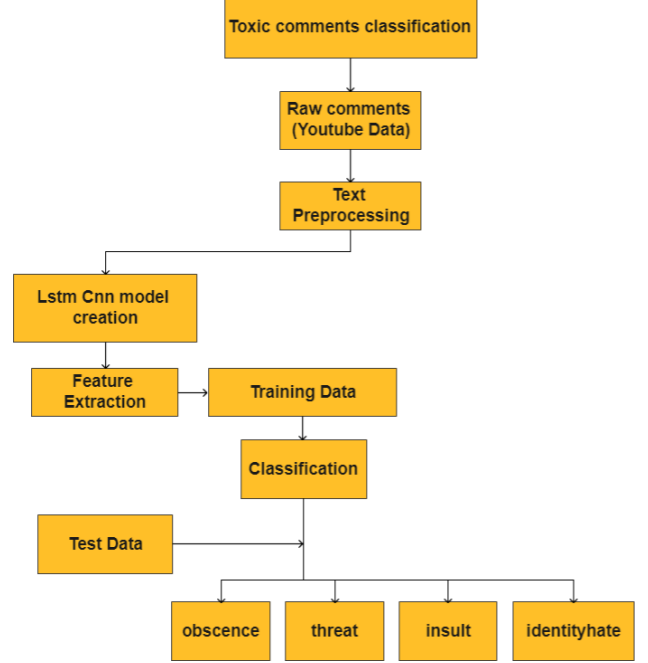

**FIGURE 1.** The flow of the proposed methodology.

Owing to the higher difference in the number of samples for toxic and non-toxic classes, various re-sampling approaches area applied. Random undersampling improve the performance of the proposed methodology. The ratio of the number of samples after re-sampling is given in Table 3.

**TABLE 3.** Number of samples after applying re-sampling.

| Category | Count | Exp.Data | Under-sampling | Over-sampling |
|---|---|---|---|---|
| Non-Toxic | 143346 | 70000 | 15294 | 70000 |
| Toxic | 15294 | 15294 | 15294 | 70000 |
| Total | 158640 | 85294 | 30588 | 140000 |

In under-sampling random samples of the majority class are removed while in over-sampling, the samples of the minority class are generated After re-sampling, the data is split into training and testing sets with a 75:25 ratio. The number of training and testing samples after data split is given in Table 4.

**TABLE 4.** Number of samples for train and test data.

| Re-sampling | Set | Toxic | Non-Toxic | Total |
|---|---|---|---|---|
| Without re-sampling | Training | 11520 | 52450 | 63970 |
|  | Testing | 3774 | 17550 | 21324 |
|  | Total | 15294 | 70000 | 85294 |
| After Under-sampling | Training | 11511 | 11430 | 22941 |
|  | Testing | 3783 | 3864 | 7647 |
|  | Total | 15294 | 15294 | 30588 |
| After Over-sampling | Training | 52342 | 52658 | 105000 |
|  | Testing | 17658 | 17342 | 35000 |
|  | Total | 70000 | 70000 | 140000 |

We used the training set to train the machine learning models and the proposed ensemble classifier on extracted features and evaluate the performance of machine learning models on the test data. For performance evaluation, various metrics including accuracy, precision, recall, and F1 score are used.

### H. SUPERVISED MACHINE LEARNING MODELS

Various machine learning models are adopted to perform toxic comments classification. Machine learning algorithms are implemented using the Scikit-learn library. We used two tree-based models such as RF, GBM, two linear models LR, SVM, and one non-parametric model KNN. The hyper-parameters of all the machine learning models are given in Table 5.

**TABLE 5.** Machine learning models parameters.

| Algorithm | Hyper parameters |
|---|---|
| RF | n_estimators=300, random_state=5, max_depth=100 |
| GBM | n_estimators=100, max_depth=100 |
| LR | C=1.0, max_iter=100, penalty='l2' |
| SVM | kernel='linear', C=2.0, random_state=500 |
| KNN | algorithm='auto', leaf_size=30, n_neighbors=3 |

#### 1) SUPPORT VECTOR MACHINE

SVM is a supervised machine learning model used for both classification and regression problems. The straightforward approach to classifying the data starts by constructing a function that divides the data points into consistent labels with (a) the least amount of errors possible or (b) the highest possible margin. That is because larger empty areas next to the splitting function contribute to fewer errors. After the function is constructed, the labels are better separated from each other. Hyperparameters of SVM are listed in Table 5 in which the kernel ='linear' specifies the kernel type used for SVM. The linear kernel is used to ensure high accuracy and reduced time complexity. The term C = 2.0 is used as the regularization parameter and the strength of the regularization is inversely proportional to C. The parameter random_state = 500 is used for the seed of the pseudo-random number which is used for likelihood calculations when shuffling the results.

#### 2) RANDOM FOREST

RF is a tree-based ensemble classifier, which generates predictions that are extremely accurate by combining several poor apprentices (weak learners). RF uses bootstrap bagging to train a variety of decision trees using various bootstrap samples. In RF, a bootstrap sample is produced by subsampling the training data set, where the size of a sample dataset and the training dataset sample are the same. RF and other ensemble classifiers utilize decision trees for the prediction using the decision trees. The identification of the attribute for the root node at each stage is a major challenge for constructing the decision trees.

$$p = \text{mode}\{T_1(y), T_2(y), \ldots, T_m(y)\} \quad (10)$$

$$p = \text{mode}\{\sum_{m=1}^{m} [T_m(y)]\} \quad (11)$$

where $p$ is the final decision of the decision trees by majority vote, while $T_1(y)$, $T_2(y)$, $T_3(y)$, and $T_m(y)$ are the number of decision trees involved in the prediction process.

To improve the accuracy, RF was implemented with $n$ as 100 which indicates the number of trees that contribute to the prediction in an RF. The 'max_depth' is set to 60 which shows the every decision tree can go to a maximum depth of 60 levels. By specifying the depth point, the 'max_depth' parameter decreases uncertainty in the decision tree and decreases the probability of the decision tree over-fitting. The parameter 'random state' is used for the randomness of the samples during the training. For our experiments, we attain good results with RF by using only two hyperparameters.

#### 3) GRADIENT BOOSTING MACHINE

Gradient boosting classifiers is a collection of algorithms for machine learning that combine several weak learners to construct a strong prediction model. A loss function relies on the GBM and a customized loss function can also be used. The GBM supports several generic loss functions, but the loss mechanism has to be differentiable. Classification algorithms also use logarithmic loss, while squared errors can be used in regression algorithms. Every time the boosting algorithm is implemented, the gradient boosting system does not need to derive a new loss function, rather any differentiable loss function can be applied to the system. Several hyperparameters are tuned to get good accuracy from the GBM. For example, $n$ is set to 100 indicating the number of trees which contribute to the prediction. Equipped with 100 decision trees, the final prediction is made by voting all predictions of the decision trees. Value of 'max depth is used 60 allowing a decision tree to a maximum depth of 60 levels.

#### 4) LOGISTIC REGRESSION

Logistic regression is one of the most widely used approaches for binary classification problems. LR is known for the method that it uses, i.e., the logistic equation also called the sigmoid function. The sigmoid function is an S-shaped curve that can take any evaluated number and maps it to a value between 0 and 1 [50].

$$y = \frac{1}{1 + e^{-value}} \quad (12)$$

where $e$ is the base of the normal logarithms and value is the real numerical value that is to be converted. Below is a plot of numbers between -5 and 5, transformed by the logistic function into ranges 0 and 1.

$$y = \frac{e^{(b0+b1*x)}}{(1 + e^{(b0+b1*x)})} \quad (13)$$

where $b0$ is the bias or intercept, $y$ is the expected performance and $b1$ is the coefficient for the single input value $x$. Every column of the input data has a coefficient $b$ correlated with it (a constant actual value) to be learned from the training data.

To attain high accuracy, LR is used with 100 'max_iter' for the solvers to converge. The parameter 'penalty' is set to 'l2' which is used to specify the norm used in the penalization. The parameter C = 1.0 is used to specify the inverse of the regularization strength.

### 5) K-NEAREST NEIGHBOR

KNN is one of the simplest supervised classification methods in machine learning. The KNN identifies the similarities between the new data and existing cases and puts the new data in the group with high similarity. The similarity is calculated using distance calculation between the new data and the existing classes. For distance measurement, various distance estimation methods are used such as Euclidean, Manhattan, and Cityblock, etc. KNN algorithm can be used for both regression and classification, but it is mainly used for classification problems. KNN is a non-parametric algorithm, implying that it considered no inference to the underlying data. KNN has multiple parameters that can be refined to achieve high accuracy. For the current study, leaf size is set to 30 which is passed to the ball tree or KD Tree. The optimal value depends on the nature of the problem. Minkowski is used as the distance metric while the number of the neighbor is set to 3.

### 6) Long short term memory (LSTM)

LSTM networks were designed specifically to overcome the long-term dependency problem faced by recurrent neural networks RNNs LSTMs have feed*back* connections which make them different to more traditional feed*forward* neural networks. This property enables LSTMs to process entire sequences of data (e.g. time series) without treating each point in the sequence independently, but rather, retaining useful information about previous data in the sequence to help with the processing of new data points. As a result, LSTMs are particularly good at processing sequences of data such as text, speech and general time-series.

### 7) Convolutional Neural networks (CNN)

The construction of a convolutional neural network is a multi-layered feed-forward neural network, made by assembling many unseen layers on top of each other in a particular order.
It is the sequential design that give permission to CNN to learn hierarchical attributes.
In CNN, some of them followed by grouping layers and hidden layers are typically convolutional layers followed by activation layers.

*I. EVALUATION METRICS*

We evaluate the performance of machine learning models in terms of accuracy, precision, recall, and F1 score.

### 1) ACCURACY

Accuracy indicates the ratio of correct predictions to the total predictions from the classifiers on test data. The maximum accuracy score is 1 indicating that all predictions from the classifier are correct while the minimum accuracy score can be 0. Accuracy can be calculated as

$$Accuracy = \frac{Number\ of\ correct\ predictions}{Total\ number\ of\ predictions}, \quad (14)$$

Another form to calculated accuracy is using

$$Accuracy = \frac{TP+TN}{TP+FP+TN+FN} \quad (15)$$

where TP is a true positive, TN is a true negative, FP is a false positive and FN is a false negative.

### 2) PRECISION

Precision is also known as a positive predictive value and represents the relative number of correctly classified instances among all true classified instances. A precision value of 1 means that every instance of data that is categorized as positive which is positive. It is important to note, however, that this does not influence the number of positive instances with the label negative which are predicted as positive.

$$Precision = \frac{TP}{TP+FP} \quad (16)$$

3) RECALL

Recall often called sensitivity represents the relative number of positive classified instances from all positive instances. The recall is defined as

$$Recall = \frac{TP}{TP+FN} \quad (17)$$

4) F1 SCORE

Precision and recall are not regarded as true representers of the performance of a classifier individually. F1 has been deemed more important as it combines both precision and recall and gives a score between 0 and 1. It is the harmonic mean of precision and recall and calculated using

$$F_1 = 2 \times \frac{Precision \times Recall}{Precision + Recall} \quad (18)$$

## IV. RESULTS AND DISCUSSION

Several experiments are performed to evaluate the performance of both the selected machine learning classifiers, as well as, the proposed RVVC ensemble classifier. The experiments are divided into three categories: experiments without re-sampling, experiments with under-sampling, and experiments with over-sampling.

### A. PERFORMANCE OF MACHINE LEARNING MODELS ON IMBALANCED DATASET

Initial experiments are performed using the original imbalanced dataset with TF-IDF and BoW separately. Tables 6 shows the values of performance evaluation metrics on the imbalanced dataset using TF-IDF features. There is a lot of fluctuation in the values of evaluation parameters. For example, the accuracy of RF using TF-IDF is 0.92 but the F1 score is 0.83. The difference in the values of accuracy and F1 score is similar for other machine learning models.

TABLE 6. Performance results of all models on imbalanced dataset using TF-IDF.

| Classifier | Accuracy | Precision |
|---|---|---|
| RF | 0.92 | 0.94 |
| SVC | 0.94 | 0.90 |
| KNN | 0.86 | 0.89 |
| DT | 0.91 | 0.85 |
| LSTM | 0.96 | 0.91 |
| CNN | 0.94 | 0.87 |
| Lstm -Cnn | 0.977 | 0.89 |

Table 7 shows the results for machine learning models when trained and tested using the BoW features on the imbalanced dataset. Results indicate that the models get over-fitted on the majority class data because the models get more data from the majority class as compared to the minority class. Consequently, the number of wrong predictions for the minority class is higher than the majority class.

Owing to the high difference in the values of accuracy and F1 score, correct predictions (CP) and wrong predictions (WP) are important evaluation parameters to be analyzed. Table 8 shows the TP, TN, FP, CP, and WP for both TF-IDF and BoW for all the classifiers. Results show

TABLE 7. Performance results of all models on imbalanced dataset using BoW.

| Classifier | Accuracy | Precision |
|---|---|---|
| RF | 0.92 | 0.94 |
| SVC | 0.94 | 0.90 |
| KNN | 0.86 | 0.89 |
| DT | 0.91 | 0.85 |
| LSTM | 0.96 | 0.91 |
| CNN | 0.94 | 0.87 |
| Lstm - Cnn | 0.977 | 0.89 |

### B. PERFORMANCE OF MODELS ON BALANCED DATASET USING UNDER-SAMPLING

Further experiments are performed using a balanced dataset with a random under-sampling technique. Results using TF-IDF features on the under-sampled data are shown in Results suggest that the performance of the selected models has been degraded on the under-sampled dataset. As under-sampling reduces the size of the dataset, the number of features to train the models is also reduced which affects the accuracy of machine learning models. It is observed that the difference in the values of accuracy and other evaluation parameters has been reduced and the values foraccuracyandF1aresimilarnow.It indicates the good fit ofmachine learning models. Lstm-cnn model outperforms all other models in terms of accuracy, precision, recall, and F1 score when used with TF-IDF feature from the under-sampled dataset. It achieves the highest accuracy and F1 with a value of 0.91 each and performs better than all other classifiers.Table 10 shows the performance of machine learning models after under-sampling with the features. This performance shows that the ensemble model can also perform well on a small dataset resulting from the under-sampling. The proposed Lstm-Cnn model is a combination of lstm and cnn

## C. EXPERIMENT RESULTS OF MODELS USING OVER-SAMPLED DATA

Experiments are performed using the balanced dataset both with TF-IDF features. Table 12 shows the performance of all models with TF-IDF features.

**TABLE 12.** Performance results of all models using TF-IDF features from over-sampled dataset.

| Classifier | Accuracy | Precision |
|---|---|---|
| RF | 0.92 | 0.94 |
| SVC | 0.94 | 0.90 |
| KNN | 0.86 | 0.89 |
| DT | 0.91 | 0.85 |
| LSTM | 0.96 | 0.91 |
| CNN | 0.94 | 0.87 |
| Lstm - Cnn | 0.977 | 0.89 |

The performance of machine learning models has improved significantly when trained on TF-IDF features from the over-sampled dataset. Over-sampling increases the dataset size which increases the number of features for training the models. Consequently, it helps to a good fit of models and increases their performance. However, at the same time the performance of the KNN model, which does not perform well on large features set, has degraded. Among all models, Lstm-cnn achieves the highest accuracy of 0.97 and outperforms all other models when trained on TF-IDF features.Table 13 shows the performance of the models with BoW features from the over-sampled dataset. Lstm - cnn performs well features as well and achieves a joint accuracy of 0.93 with LR. However, its recall score is higher than Lstm which shows its superior performance. KNN models show poor performance among all the classifiers with an accuracy of 0.64 while RF and SVC perform well with 0.90 and 0.91 accuracies, respectively. As a whole, the performance of all the models has been reduced with BoW features than that of TF-IDF features.

**TABLE 13.** Performance results of all models on over-sampled data using BoW features.

| Classifier | Accuracy | Precision |
|---|---|---|
| RF | 0.92 | 0.94 |
| SVC | 0.94 | 0.90 |
| KNN | 0.86 | 0.89 |
| DT | 0.91 | 0.85 |
| LSTM | 0.96 | 0.91 |
| CNN | 0.94 | 0.87 |
| Lstm - Cnn | 0.977 | 0.89 |

## E. PERFORMANCE ANALYSIS WITH STATE-OF-THE-ART APPROACHES

Performance comparison of the proposed Lstm-cnn is done with five state-of-the-art approaches including both machine and deep learning approaches for toxic comments classification. Table 16 shows the performance appraisal results for Lstm-Cnn and other models. Results prove that the proposed Lstm-Cnn performs better than other approaches to correctly classify the toxic and non-toxic comments.

## F. STATISTICAL T-TEST

To show the significance of the proposed Lstm-Cnn model, a statistical significance test, a T-test has been performed. To support the T-test we suppose two hypotheses as follow:
- Null hypotheses: The proposed model Lstm-cnn is statistically significant.
- Alternative hypotheses: The Proposed model is not statistically significant.

Statistical T-test results that the Lstm-cnn is statistically significant for all resampling cases. Lstm-cnn accepts null hypotheses without resampling, under-sampling, and oversampling.

## V. CONCLUSION

This study analyzes the performance of various machine learning models to perform toxic comments classification and proposes an ensemble approached called Lstm-cnn. The influence of an imbalanced dataset and balanced datasetusing random under-sampling and over-sampling on the performance of the models is analyzed through extensive experiments. Two feature extraction approaches including TF-IDF a are used to get the feature vector for models' training. Results indicate that models perform poorly on the imbalanced dataset while the balanced dataset tends to increase the classification accuracy. Besides the machine learning classifiers like SVM, RF, GBM, and LR, the proposed RVVC and RNN deep learning models perform well with the balanced dataset. The performance with an over-sampled dataset is better than the under-sampled dataset as the feature set is large when the data is over-sampled which elevates the performance of the models. Results suggest that balancing the data reduces the chances of models over-fitting which happens if the imbalanced dataset is used for training. Moreover, TF-IDF shows better classification accuracy for toxic comments The proposed ensemble approach Lstm-cnn demonstrates its efficiency for toxic and non-toxic comments classification. The performance of Lstm-cnn is superior both with the imbalanced and balanced dataset, yet, it achieves the highest accuracy of 0.97 when used with TF-IDF features The performance comparisonwith state-of-the-art approaches also indicates that Lstm-cnn shows better performance and proves good on small and large feature vectors. Despite the better performance of the proposed ensemble approach, its computational complexity is higher than the individual models which is an important topic for our future research. Similarly, dataset imbalance can overstate the results because data balancing using or random under-sampling approach may have a certain influence on the reported accuracy. Moreover, we intend to perform further experiments on multi-domain datasets and run experiments on more datasets for toxic comment classification.

## REFERENCES

[1] E. Aboujaoude, M. W. Savage, V. Starcevic, and W. O. Salame, ''Cyberbullying: Review of an old problem gone viral,'' *J. Adolescent Health*, vol. 57, no. 1, pp. 10–18, Jul. 2015.

[2] *How Much Data is Created on the Internet Each Day?* Accessed: Jun. 6, 2020. [Online]. Available: https://blog.microfocus.com/how-muchdata-is-created-on-the-internet-each-day/

[3] *World Internet Users and 2020 Population Stats.* Accessed: Jun. 6, 2020. [Online]. Available: https://www.internetworldstats.com/stats.htm

[4] M. Duggan, ''Online harassment,'' Pew Res. Center, Washington, DC, USA, Tech. Rep., 2014. [Online]. Available: https://www.pewresearch.org/internet/wp-content/uploads/sites/9/2017/07/PI_2017.07.11_OnlineHarassment_FINAL.pdf

[5] P. Badjatiya, S. Gupta, M. Gupta, and V. Varma, ''Deep learning for hate speech detection in tweets,'' in *Proc. 26th Int. Conf. World Wide Web Companion*, 2017, pp. 759–760.

[6] *Man Jailed for 35 years in Thailand for Insulting Monarchy on Facebook.* Accessed: Jun. 6, 2020. [Online]. Available: https://www.theguardian.com/world/2017/jun/09/man-jailed-for-35-years-in-thailand-forinsulting-monarchy-on-facebook

[7] *Mississippi Teacher Fired After Racist Facebook Post; Black Parent Responds.* Accessed: Jun. 6, 2020. [Online]. Available: https://www.clarionledger.com/story/news/2017/09/20/mississippi-teacher-fired-afterracist-facebook-post/684264001/

[8] E. Wulczyn, N. Thain, and L. Dixon, ''Ex machina: Personal attacks seen at scale,'' in *Proc.26thInt.Conf.WorldWideWeb*, Apr.2017, pp. 1391–1399.

[9] M. Ptaszynski, J. K. K. Eronen, and F. Masui, ''Learning deep on cyberbullying is always better than brute force,'' in *Proc. LaCATODA@ IJCAI*, 2017, pp. 3–10.

[10] S. Agrawal and A. Awekar, ''Deep learning for detecting cyberbullying across multiple social media platforms,'' in *Proc. Eur. Conf. Inf. Retr.* Cham, Switzerland: Springer, 2018, pp. 141–153.

[11] M. Ibrahim, M. Torki, and N. El-Makky, ''Imbalanced toxic comments classification using data augmentation and deep learning,'' in *Proc. 17th IEEE Int. Conf. Mach. Learn. Appl. (ICMLA)*, Dec. 2018, pp. 875–878.

[12] M. A. Saif, A. N. Medvedev, M. A. Medvedev, and T. Atanasova, ''Classification of online toxic comments using the logistic regression and neural networks models,'' *AIP Conf. Proc.*, vol. 2048, no. 1, 2018, Art. no. 060011.

[13] S.V.Georgakopoulos,S.K.Tasoulis,A.G.Vrahatis,andV.P.Plagianakos, ''Convolutional neural networks for toxic comment classification,'' in *Proc. 10th Hellenic Conf. Artif. Intell.*, Jul. 2018, pp. 1–6.

[14] S. Zaheri, J. Leath, and D. Stroud, ''Toxic comment classification,'' *SMU Data Sci. Rev.*, vol. 3, no. 1, p. 13, 2020.

[15] R. Beniwal and A. Maurya, ''Toxic comment classification using hybrid deep learning model,'' in *Sustainable Communication Networks and Application*. Cham, Switzerland: Springer, 2021, pp. 461–473.

[16] A. N. M. Jubaer, A. Sayem, and M. A. Rahman, ''Bangla toxic comment classification (machine learning and deep learning approach),'' in *Proc. 8th Int.Conf.Syst.ModelingAdv.Res.Trends(SMART)*, Nov.2019, pp. 62–66.

[17] B. van Aken, J. Risch, R. Krestel, and A. Löser, ''Challenges for toxic comment classification: An in-depth error analysis,'' 2018, *arXiv:1809.07572*. [Online]. Available: http://arxiv.org/abs/1809.07572

[18] H. H. Saeed, K. Shahzad, and F. Kamiran, ''Overlapping toxic sentiment classification using deep neural architectures,'' in *Proc. IEEE Int. Conf. Data Mining Workshops (ICDMW)*, Nov. 2018, pp. 1361–1366.

[19] S. Malmasi and M. Zampieri, ''Detecting hate speech in social media,'' 2017, *arXiv:1712.06427*. [Online]. Available: http://arxiv.org/abs/1712.06427

[20] Z. Zhang, D. Robinson, and J. Tepper, ''Detecting hate speech on Twitter using a convolution-GRU based deep neural network,'' in *Proc. Eur. Semantic Web Conf.* Cham, Switzerland: Springer, 2018, pp. 745–760.

[21] P. A. Ozoh, M. O. Olayiwola, and A. A. Adigun, ''Identification and classification of toxic comments on social media using machine learning techniques,'' *Int. J. Res. Innov. Appl. Sci.*, vol. 4, no. 9, pp. 1–6, Nov. 2019.

[22] S. Carta, A. Corriga, R. Mulas, D. Recupero, and R. Saia, ''A supervised multi-class multi-label word embeddings approach for toxic comment classification,'' in *Proc. KDIR*, 2019, pp. 105–112.

[23] B. Gambäck and U. K. Sikdar, ''Using convolutional neural networks to classify hate-speech,'' in *Proc. 1st Workshop Abusive Lang. Online*, 2017, pp. 85–90.

[24] S. R. Basha, J. K. Rani, J. P. Yadav, and G. R. Kumar, ''Impact of feature selection techniques in text classification: An experimental study,'' *J. Mech. Continua Math. Sci.*, no. 3, pp. 39–51, 2019.

[25] S. R. Basha and J. K. Rani, ''A comparative approach of dimensionality reduction techniques in text classification,'' *Eng., Technol. Appl. Sci. Res.*, vol. 9, no. 6, pp. 4974–4979, Dec. 2019.

[26] M. S. Basha, S. K. Mouleeswaran, and K. R. Prasad, ''Sampling-based visual assessment computing techniques for an efficient social data clustering,'' *J. Supercomput.*, pp. 1–25, Jan. 2021.

[27] C. Nobata, J. Tetreault, A. Thomas, Y. Mehdad, and Y. Chang, ''Abusive language detection in online user content,'' in *Proc. 25th Int. Conf. World Wide Web*. Republic and Canton of Geneva, Switzerland: International World Wide Web Conferences Steering Committee, Apr. 2016, pp. 145–153.

[28] R. Martins, M. Gomes, J. J. Almeida, P. Novais, and P. Henriques, ''Hate speech classification in social media using emotional analysis,'' in *Proc. 7th Brazilian Conf. Intell. Syst. (BRACIS)*, Oct. 2018, pp. 61–66.

[29] H. Hosseini, S. Kannan, B. Zhang, and R. Poovendran, ''Deceiving Google's perspective API built for detecting toxic comments,'' 2017, *arXiv:1702.08138*. [Online]. Available: http://arxiv.org/abs/1702.08138

[30] *Toxic Comment Classification Challenge*. Accessed: May 5, 2020. [Online]. Available: https://www.kaggle.com/c/jigsaw-toxic-commentclassification-challenge

[31] S. Alam and N. Yao, ''The impact of preprocessing steps on the accuracy of machine learning algorithms in sentiment analysis,'' *Comput. Math. Org. Theory*, vol. 25, no. 3, pp. 319–335, Sep. 2019.

[32] F. Rustam, I. Ashraf, A. Mehmood, S. Ullah, and G. Choi, ''Tweets classification on the base of sentiments for US airline companies,'' *Entropy*, vol. 21, no. 11, p. 1078, Nov. 2019.

[33] M. Anandarajan, C. Hill, and T. Nolan, *Practical Text Analytics: Maximizing the Value of Text Data* (Advances Analytics Data Science), vol. 2. Cham, Switzerland: Springer, 2019.

[34] Z. Z. Wint, T. Ducros, and M. Aritsugi, ''Spell corrector to social media datasets in message filtering systems,'' in *Proc. 12th Int. Conf. Digit. Inf. Manage. (ICDIM)*, Sep. 2017, pp. 209–215.

[35] S. Yang and H. Zhang, ''Text mining of Twitter data using a latent Dirichlet allocation topic model and sentiment analysis,'' *Int. J. Comput. Inf. Eng*, vol. 12, no. 7, pp. 525–529, 2018.

[36] F. F. Bocca and L. H. A. Rodrigues, ''The effect of tuning, feature engineering, and feature selection in data mining applied to rainfed sugarcane yield modelling,'' *Comput. Electron. Agricult.*, vol. 128, pp. 67–76, Oct. 2016.

[37] J. Heaton, ''An empirical analysis of feature engineering for predictive modeling,'' in *Proc. SoutheastCon*, Mar. 2016, pp. 1–6.

[38] S. C. Eshan and M. S. Hasan, ''An application of machine learning to detect abusive Bengali text,'' in *Proc. 20th Int. Conf. Comput. Inf. Technol. (ICCIT)*, Dec. 2017, pp. 1–6.

[39] F. Rustam, M. Khalid, W. Aslam, V. Rupapara, A. Mehmood, and G. S. Choi, ''A performance comparison of supervised machine learning models for COVID-19 tweets sentiment analysis,'' *PLoS ONE*, vol. 16, no. 2, Feb. 2021, Art. no. e0245909.

[40] E. B. Fatima, B. Omar, E. M. Abdelmajid, F. Rustam, A. Mehmood, and G. S. Choi, ''Minimizing the overlapping degree to improve classimbalanced learning under sparse feature selection: Application to fraud detection,'' *IEEE Access*, vol. 9, pp. 28101–28110, 2021.

[41] T. Elhassan and M. Aljurf, ''Classification of imbalance data using Tomek link (T-link) combined with random under-sampling (RUS) as a data reductionmethod,''*GlobalJ.Technol.Optim.*,vol.1,no.1,pp. 1–11,2016.

[42] J. Prusa, T. M. Khoshgoftaar, D. J. Dittman, and A. Napolitano, ''Using random undersampling to alleviate class imbalance on tweet sentiment data,'' in *Proc. IEEE Int. Conf. Inf. Reuse Integr.*, Aug. 2015, pp. 197–202.

[43] W.-C. Lin, C.-F. Tsai, Y.-H. Hu, and J.-S. Jhang, ''Clustering-based undersampling in class-imbalanced data,'' *Inf. Sci.*, vols. 409–410, pp. 17–26, Oct. 2017.

[44] A. R. Hassan and M. I. H. Bhuiyan, ''Automated identification of sleep states from EEG signals by means of ensemble empirical mode decomposition and random under sampling boosting,'' *Comput. Methods Programs Biomed.*, vol. 140, pp. 201–210, Mar. 2017.

[45] N. V. Chawla, K. W. Bowyer, L. O. Hall, and W. P. Kegelmeyer, ''SMOTE: Synthetic minority over-sampling technique,'' *J. Artif. Intell. Res.*, vol. 16, pp. 321–357, Jun. 2002.

[46] D. J. Dittman, T. M. Khoshgoftaar, R. Wald, and A. Napolitano, ''Comparison of data sampling approaches for imbalanced bioinformatics data,'' in *Proc. 27th Int. FLAIRS Conf.*, 2014, pp. 1–4.

[47] R. Blagus and L. Lusa, ''Smote for high-dimensional class-imbalanced data,'' *BMC Bioinf.*, vol. 14, Mar. 2013, Art. no. 106.

[48] A. Fernández, S. Garcia, F. Herrera, and N. V. Chawla, ''SMOTE for learning from imbalanced data: Progress and challenges, marking the 15year anniversary,'' *J. Artif. Intell. Res.*, vol. 61, pp. 863–905, Apr. 2018.

[49] R. Blagus and L. Lusa, ''Evaluation of SMOTE for high-dimensional classimbalanced microarray data,'' in *Proc. 11th Int. Conf. Mach. Learn. Appl.*, Dec. 2012, pp. 89–94.

[50] GeeksforGeeks. *Ml | Cost Function in Logistic Regression*. Accessed: Dec. 16, 2019. [Online]. Available: https://www.geeksforgeeks.org/mlcost-function-in-logistic-regression/

[51] M. Umer, I. Ashraf, A. Mehmood, S. Kumari, S. Ullah, and G. S. Choi, ''Sentiment analysis of tweets using a unified convolutional neural network-long short-term memory network model,'' *Comput. Intell.*, vol. 37, no. 1, pp. 409–434, Feb. 2021.

. . .